\documentclass[
]{ceurart}

\usepackage{listings}
\usepackage{pifont}

\usepackage{xspace}

\makeatletter
\DeclareRobustCommand\onedot{\futurelet\@let@token\@onedot}
\def\@onedot{\ifx\@let@token.\else.\null\fi\xspace}

\def\eg{\emph{e.g}\onedot}

\def\etal{\emph{et al}\onedot}
\makeatother

\begin{document}

\copyrightyear{2025}
\copyrightclause{Copyright for this paper by its authors. Use permitted under Creative Commons License Attribution 4.0 International (CC BY 4.0).}

\conference{MiGA@IJCAI25: International IJCAI Workshop on 3rd Human Behavior Analysis for Emotion Understanding, August 29, 2025, Guangzhou, China.}

\title{MM-Gesture: Towards Precise Micro-Gesture Recognition through Multimodal Fusion}


\author[1]{Jihao Gu}[%
orcid=0009-0009-0141-4807,
email=jihao.gu.23@ucl.ac.uk,
]

\author[2,5,6]{Fei Wang}[%
orcid=0009-0004-1142-6434,
email=jiafei127@gmail.com,
]

\author[3]{Kun Li}[%
orcid=0000-0001-5083-2145,
email=kunli.hfut@gmail.com,
]
\cormark[1]

\author[2,4]{Yanyan Wei}[%
orcid=0000-0001-8818-6740,
email=weiyy@hfut.edu.cn,
]

\author[3]{Zhiliang Wu}[%
orcid=0000-0002-6597-8048,
email=wu_zhiliang@zju.edu.cn,
]

\author[2,5]{Dan Guo}[%
orcid=0000-0003-2594-254X,
email=guodan@hfut.edu.cn,
]

\address[1]{University College London (UCL), Gower Street, London, WC1E 6BT, UK}
\address[2]{School of Computer Science and Information Engineering, School of Artificial Intelligence, Hefei University of Technology (HFUT)}
\address[3]{ReLER, CCAI, Zhejiang University, China}
\address[4]{Key Laboratory of Knowledge Engineering with Big Data (HFUT), Ministry of Education}
\address[5]{Institute of Artificial Intelligence, Hefei Comprehensive National Science Center, China}
\address[6]{Xinsight Lab, Research Institute, Hefei Zhongjuyuan Intelligent Technology Co., Ltd., China}
\cortext[1]{Corresponding author.}

\begin{abstract}
In this paper, we present MM-Gesture, the solution developed by our team HFUT-VUT, which ranked 1st in the micro-gesture classification track of the 3rd MiGA Challenge at IJCAI 2025, achieving superior performance compared to previous state-of-the-art methods. MM-Gesture is a multimodal fusion framework designed specifically for recognizing subtle and short-duration micro-gestures (MGs), integrating complementary cues from joint, limb, RGB video, Taylor-series video, optical-flow video, and depth video modalities. Utilizing PoseConv3D and Video Swin Transformer architectures with a novel modality-weighted ensemble strategy, our method further enhances RGB modality performance through transfer learning pre-trained on the larger MA-52 dataset. Extensive experiments on the iMiGUE benchmark, including ablation studies across different modalities, validate the effectiveness of our proposed approach, achieving a top-1 accuracy of 73.213\%.
Code is available at: \href{https://github.com/momiji-bit/MM-Gesture}{https://github.com/momiji-bit/MM-Gesture}. 
\end{abstract}


\maketitle

\section{Introduction}
Micro‑Gestures (MGs)~\cite{chen2019analyze,liu2021imigue,chen2023smg,chen20242nd}, defined as spontaneous and fine‑grained movements, such as nose touching, hair scratching, or subtle finger rubs, encode rich affective and cognitive cues that rarely surface in conventional action recognition benchmarks. Compared with conventional actions~\cite{li2021proposal,wang2025exploiting,balazia2022bodily,li2025deemo}, MGs are unintentional, short‑duration, and confined to small body regions, which makes them extremely difficult to capture and classify. 

Due to the subtle changes and short duration of MGs, relying solely on a single modality (\eg, RGB~\cite{liu2022video,li2025prototypical,guo2024benchmarking}, skeleton~\cite{li2023joint,huang2023micro}) often captures merely partial characteristics of MGs, thus failing to fully and thoroughly exploit the comprehensive information latent in available data. 
Despite significant advancements in previous studies on micro-gesture and micro-action recognition~\cite{guo2024benchmarking,li2025prototypical,li2024mmad,li2023data,gu2025motion,sun2023unified,dong2023hierarchical}, most existing approaches remain confined to utilizing limited modalities, such as RGB combined with skeleton data~\cite{chen2024prototype,huang2024multi,duan2022revisiting}. 
However, these methods have not sufficiently leveraged the abundant and complementary information conveyed by multi-modal.

In this work, we propose a novel multi-modal fusion framework \textbf{MM‑Gesture} tailored explicitly for the challenging task of MGs classification. Specifically, we construct baseline models leveraging PoseConv3D~\cite{duan2022revisiting} and Video Swin Transformer~\cite{liu2022video,wang2025exploiting}, integrating information across six complementary modalities: \textbf{joint, limb, RGB video, Taylor video, optical flow video, and depth video}. In addition, to enhance the performance of the RGB modality, we apply transfer learning by pre-training on the Micro-Action 52 dataset~\cite{guo2024benchmarking} and fine-tuning on the iMiGUE dataset~\cite{liu2021imigue}. 

The key contributions of this paper can be summarized as follows:
\begin{itemize}
\item We present an integrated multi-modal MGs classification network that utilizes complementary information from six diverse modalities: joint, limb, RGB video, Taylor video, optical flow video, and depth video. 
\item We propose an effective ensemble fusion method capable of efficiently integrating six modalities, enabling the joint exploitation of modality-specific strengths for improved MGs classification accuracy.
\item Extensive experiments on the iMiGUE dataset~\cite{liu2021imigue} demonstrate that the proposed MM‑Gesture achieves state-of-the-art performance, reaching a Top-1 accuracy of 73.213\%, which is the highest reported accuracy in previous Micro-gesture Analysis (MiGA) challenges.
\end{itemize}

\section{Related Work}
Micro-Gestures (MGs) are becoming increasingly important in understanding human emotions, focusing on subtle body movements in daily interactions. Advances in this field have been driven by the development of large benchmark datasets and sophisticated model architectures~\cite{chen2019analyze,liu2021imigue,chen2023smg,guo2024benchmarking}.
Key datasets include the SMG dataset~\cite{chen2023smg}, which consists of recordings from 40 participants engaged in storytelling, capturing upper limb micro-gestures and emotional states. The iMiGUE dataset~\cite{liu2021imigue} offers identity-free videos of 72 athletes at press conferences, annotated with 32 micro-gesture categories for analyzing both actions and emotions. The MA-52 dataset~\cite{guo2024benchmarking} expands the focus to full-body micro-actions, with 22,000 samples covering 52 action-level and 7 body-level categories, sourced from psychological interviews to recognize subtle visual cues.

Current models primarily focus on limited modalities. RGB-based methods leverage spatial-temporal modeling strategies, such as a pure Transformer backbone with shifted 3D local attention windows~\cite{liu2022video}. 
MANet~\cite{guo2024benchmarking} integrates SE and TSM modules with semantic embedding loss for fine-grained micro-action recognition. Skeleton-based approaches include a 3D-CNN model with joint and semantic embedding losses~\cite{li2023joint}, and an EHCT framework~\cite{huang2023micro} employs hypergraph-based attention and ensemble Transformers~\cite{wang2024eulermormer,wang2024frequency} to capture high-order joint relations and address class imbalance. 
In contrast, skeleton sequences can be encoded as 3D heatmaps and fused with RGB inputs through a dual-branch multimodal network~\cite{duan2022revisiting}. Inspired by this network, 
Chen~\etal~\cite{chen2024prototype} adopt channel-wise cross-attention and prototype refinement to enhance feature fusion and category discrimination, while Huang~\etal~\cite{huang2024froster} design a multi-scale heterogeneous fusion network.
Recently, Li~\etal~\cite{li2025prototypical} propose a hierarchical prototype-based calibration method to resolve ambiguity in fine-grained actions. Overall, current methods only focus on the RGB or skeleton data. 

To exploit the complementarity between different multimodal data, we propose the MM-Gesture model, adopting a comprehensive multimodal approach that integrates six modalities: joint, limb, RGB video, Taylor video, optical flow video, and depth video. This approach enables a deeper understanding and representation of micro-gestures, capturing their nuances and dynamics. Additionally, we leverage transfer learning from the MA-52 dataset to infuse valuable prior knowledge into the RGB modality, further enhancing its recognition accuracy. Consequently, our model improves performance on existing benchmarks and paves the way for advanced applications in human emotion understanding through micro-gesture analysis.

\section{Methodology}

\begin{figure}[t!]
\centering
\includegraphics[width=\linewidth]{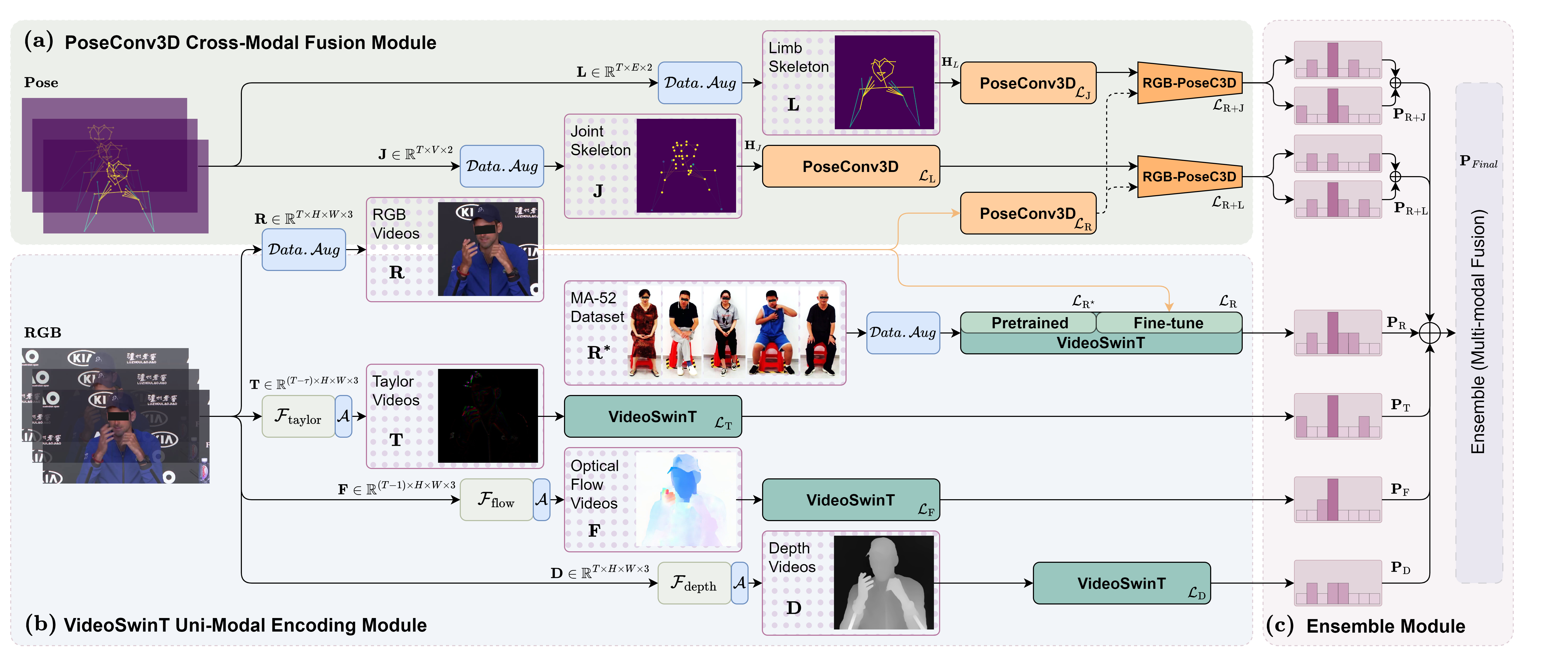}
\caption{Pipeline of the proposed multimodal micro-gesture recognition framework (\textbf{MM-Gesture}), which consists of three key components: (a) Cross-Modal Fusion Module, (b) Uni-modal Embedding Module, and (c) Ensemble Module.}
\label{fig:framework}
\end{figure}

\subsection{Data Pre-processing}

We adopted the RGB videos ($\mathbf{R} \in \mathbb{R}^{T \times H \times W \times 3}$) provided by the official dataset, along with a subset of 36 skeleton keypoints ($V$) selected from the original 137 points, to form the input joint data ($\mathbf{J} \in \mathbb{R}^{T \times V \times 2}$). These cleaned keypoints focus specifically on the upper body, hands, and facial joints. Additionally, we constructed input limb data ($\mathbf{L} \in \mathbb{R}^{T \times E \times 2}$) by computing spatial differences between adjacent joint pairs defined by the skeletal edges ($E$) connecting the selected keypoints.

To effectively capture multi-modal gesture information, we employ advanced, off-the-shelf modality extraction methods to generate complementary auxiliary modalities. Specifically, we utilize Taylor-series temporal expansion videos, optical-flow videos, and depth-estimation videos, each modality providing distinct yet complementary gesture-related information. By leveraging the ensemble among these diverse modalities, our proposed MM-Gesture model effectively exploits multi-modal feature complementarity.
\begin{equation}
\begin{aligned}
\mathbf{T} \in \mathbb{R}^{(T - \tau) \times H \times W \times 3},\quad
\mathbf{F} \in \mathbb{R}^{(T - 1) \times H \times W \times 3},\quad
\mathbf{D} \in \mathbb{R}^{T \times H \times W \times 3},\\
\mathbf{T}_t = \mathcal{F}_{\text{taylor}}(\mathbf{R}_{t:t+\tau}), \quad 
\mathbf{F}_t = \mathcal{F}_{\text{flow}}(\mathbf{R}_{t:t+1}), \quad 
\mathbf{D}_t = \mathcal{F}_{\text{depth}}(\mathbf{R}_t),
\end{aligned}
\end{equation}
\noindent{where each symbol is defined as follows:}
\begin{itemize}
\item $T$: Temporal length of the input RGB video.
\item $H, W$: Height and Width of the input RGB video frames.
\item $\tau$: Temporal window length for computing the truncated Taylor-series expansion.
\item $\mathbf{R}_t$: The RGB frame at time step $t$.
\item $\mathcal{F}_{taylor}$: The Taylor-series-based video calculated according to the approach~\cite{wang2024taylor}, where $K$ denotes the maximum order of the truncated Taylor-series expansion and $\tau$ represents the temporal window length used for aggregating local temporal context.
\item $\mathcal{F}_{flow}$: The optical-flow-based modality computed using the MemFlow network~\cite{dong2024memflow}, which estimates optical flow representations $\mathbf{F}_t$ from consecutive frames $\mathbf{R}_t$ and $\mathbf{R}_{t+1}$.
\item $\mathcal{F}_{depth}$: The depth-estimation-based modality generated using the monocular depth estimation algorithm~\cite{chen2025video}, resulting in depth representations $\mathbf{D}_t$.
\end{itemize}

\subsection{Network Architecture}

As shown in Figure~\ref{fig:framework}, the proposed multi-modal micro-gesture recognition framework (\textbf{MM-Gesture}) consists of three main modules:

\textbf{Cross-Modal Fusion Module:}
In this module, skeletal coordinates are initially transformed into Gaussian heatmap-based 3D volumes ($\mathbf{H}$) for Joint and Limb modalities individually. RGB, Joint, and Limb modalities are all separately trained through PoseConv3D~\cite{duan2022revisiting}, capturing spatial-temporal skeleton dynamics and RGB spatial context, respectively. Subsequently, the extracted RGB and skeleton features are combined via a cross-modal fusion training stage to exploit complementary information between these modalities comprehensively.

\textbf{Uni-Modal Encoding Module:}   
We leverage the VideoSwinT network~\cite{liu2022video} to independently encode four distinct modalities: RGB frames, Taylor-based temporal encoding, optical flow (computed via MemFlow), and depth estimates. Specifically, for the RGB modality, we first employ transfer learning by pretraining VideoSwinT on the MA-52 dataset and subsequently fine-tune the pretrained model on the iMiGUE dataset. For the remaining modalities (Taylor, optical flow, and depth), VideoSwinT is directly trained from scratch on the iMiGUE dataset. VideoSwinT uses a 3D shifted-window self-attention mechanism that effectively captures fine-grained spatial-temporal details within each modality.

\textbf{Ensemble Module:} 
Probabilities from the PoseConv3D Cross-Modal Fusion Module and VideoSwinT Uni-Modal Encoding Module are combined via weighted ensemble, with weights set empirically according to validation performance. This integration approach effectively exploits modality complementarity, improving robustness and accuracy in micro-gesture recognition.

\subsection{PoseConv3D Cross-Modal Fusion Module}

To effectively align skeleton-based information (consisting of joints and limbs) with RGB video representations and facilitate fine-grained complementary interactions across these modalities, we adopt PoseConv3D~\cite{duan2022revisiting} for cross-modal integration.

Specifically, we first transform the 2D coordinates of skeletal keypoints into heatmap-based representations. By applying Gaussian distributions and calculating the heatmap values using the point-to-segment distance formula, we compute and stack the heatmaps of each keypoint across all frames to generate 3D heatmap volumes. The resulting heatmaps are as follows:
\begin{equation}
\mathbf{H}_{J} \in \mathbb{R}^{T \times H \times W \times V}, \quad  \mathbf{H}_{L} \in \mathbb{R}^{T \times H \times W \times E},
\end{equation}
where $\mathbf{H}_{J}$ denotes the joint-position heatmaps, and $\mathbf{H}_{L}$ denotes the limb-connection heatmaps. Here, $T$ is the total number of frames, $V$ is the number of skeletal joints, and $E$ is the number of skeletal limbs (connections between joints). $H$, and $W$ represent the spatial resolution (height and width) of each heatmap. Subsequently, the RGB frames $\mathbf{R} \in \mathbb{R}^{T \times H \times W \times 3}$ and skeleton heatmaps $\mathbf{H}_{J}, \mathbf{H}_{L}$ are taken as input data. 

Prior to network training, data augmentation processes (\eg, scaling, cropping) are consistently applied to both RGB video frames and skeleton heatmap modalities to enhance data diversity and improve model robustness. Subsequently, the augmented data from each modality is separately forwarded into the PoseConv3D module, which extracts deep spatiotemporal feature representations. The PoseConv3D network generates modality-specific predictions denoted formally as $\hat{\mathbf{y}}_{m}$, where $ m \in \{R, J, L\} $ indicates RGB, joint heatmap, and limb heatmap modalities, respectively. Each modality-specific network is initially pretrained independently by minimizing the cross-entropy (CE) classification loss:
\begin{equation}
\mathcal{L}_{m} = \mathrm{CE}\left(\hat{\mathbf{y}}_{m},\, y\right), \quad m \in \{\text{R},\text{J},\text{L}\},
\end{equation}
where $y$ denotes the ground-truth action labels.

Next, we conduct a joint fine-tuning procedure by simultaneously optimizing combined RGB and skeleton-based modalities using the following paired-training losses:
\begin{equation}
\mathcal{L}_{\text{R+J}} = \mathcal{L}_{\text{R}} + \mathcal{L}_{\text{J}}, \quad \mathcal{L}_{\text{R+L}} = \mathcal{L}_{\text{R}} + \mathcal{L}_{\text{L}}.
\end{equation}

During model inference, the predictions yielded by distinct modalities are integrated at the probability level via a late fusion strategy. Formally, let $ P^{\star} = \operatorname{SoftMax}(\hat{\mathbf{y}}^{\star}) $, $ \star \in \{R,J,L\} $, represent modality-specific probability distributions. We then fuse predictions through average fusion to achieve final predictive distributions:
\begin{equation}
\mathbf{P}_{\text{R+J}} = \frac{1}{2}(\mathbf{P}_{\text{R}} + \mathbf{P}_{\text{J}}), \quad \mathbf{P}_{\text{R+L}} = \frac{1}{2}(\mathbf{P}_{\text{R}} + \mathbf{P}_{\text{L}}).
\end{equation}

\subsection{VideoSwinT Uni-Modal Encoding Module}

Unlike existing skeleton-video modality fusion methods, we propose a multimodal framework based on the VideoSwinT~\cite{liu2022video}, which encodes RGB video, optical flow video, Taylor-expanded video, and depth video. This encoding strategy effectively integrates color, texture, dynamic motion, and geometric structural information to better capture multidimensional micro-action features, thus enabling more fine-grained action recognition.

Specifically, we independently optimize each modality-specific backbone by minimizing the cross-entropy (CE) classification loss. Prior to training on the target iMiGUE dataset, the RGB modality network is initially pretrained on the MA-52 dataset ($\mathbf{R}^{\star}\in \mathbb{R}^{T\times H\times W\times 3}$)~\cite{guo2024benchmarking}, which provides extensive coverage of 52 types of micro-actions. After pretraining, the RGB modality network is fine-tuned on the iMiGUE dataset along with other modalities. The loss functions for pretraining and fine-tuning, along with the probability computation, are formulated as follows:
\begin{equation}
\begin{aligned}
\mathcal{L}_{m} &= \mathrm{CE}(\mathbf{\hat{y}}_m, y), \quad m\in\{\text{R}^\star,\text{R},\text{T},\text{F},\text{D}\}, \\
\mathbf{P}_m &= \operatorname{SoftMax}(\mathbf{\hat{y}}_m), \quad m \in \{\text{R},\text{T},\text{F},\text{D}\}.
\end{aligned}
\end{equation}

\subsection{Ensemble Module}

In the final ensemble stage, we introduce a probability-based weighted fusion strategy to effectively aggregate predictions derived from multiple modality-specific networks. Specifically, class probability vectors independently output by the PoseConv3D ($\mathbf{RGB + J}$, $\mathbf{RGB\!+\!L}$) and VideoSwin Transformer ($\mathbf{RGB^{*}}$, $\mathbf{Taylor}$, $\mathbf{Flow}$, $\mathbf{Depth}$) models are integrated using empirically determined weights obtained via validation-set performance.

The ensemble prediction ($\mathbf{P}_{\text{final}} \in \mathbb{R}^{cls}$) is computed by summing the weighted contributions of individual modality-specific probabilities, as follows:
\begin{equation}
\mathbf{P}_{\text{final}} = \sum_{} w_{i} \mathbf{P}_{i}, \quad  i \in \{\text{R+J}, \text{R+L}, \text{R}, \text{T}, \text{F}, \text{D}\}
\end{equation}
where each weight $w_i$ is selected based on the classification performance observed on validation samples.

This proposed ensemble-based fusion mechanism enables comprehensive exploitation of the complementary strengths inherent in multiple modality-specific models, thereby significantly improving the robustness and overall effectiveness of our multi-modal micro-gesture recognition framework.

\section{Experiments}

\subsection{Experimental Setup}
\noindent\textbf{Dataset.}
\textbf{iMiGUE} (identity-free video dataset for Micro-Gesture Understanding and Emotion analysis) dataset~\cite{liu2021imigue} consists of micro-gestures (MGs) primarily involving upper limbs, collected from post-match press conference videos of professional tennis players. It includes 31 MG categories and an additional non-MG class, comprising a total of 18,499 labeled MG samples annotated from 359 long video sequences (ranging from 0.5 to 26 minutes), totaling approximately 3.77 million frames. The dataset provides two modalities: RGB videos and corresponding 2D skeletal joint data extracted via OpenPose. iMiGUE adopts a cross-subject evaluation protocol, splitting 72 subjects into 37 for training and 35 for testing, with 12,893 samples in the training set, 777 in validation, and 4,562 in testing. 
In addition, we pretrain the proposed method on the \textbf{Micro-Action 52}~\cite{guo2024benchmarking} dataset and then fine-tune it on the iMiGUE dataset. Micro-Action 52 is a large-scale, whole-body micro-action dataset collected by a professional interviewer to capture unconscious human micro-action behaviors. The dataset contains 22,422 (22.4K) samples interviewed from 205 participants, where the annotations are categorized into two levels: 7 \textit{body-level} and 52 \textit{action-level} micro-action categories. There are 11,250, 5,586, and 5,586 instances in the training, validation, and test sets, respectively.

\noindent\textbf{Evaluation Metrics.} 
For the micro-gesture classification challenge, we employ top-1 accuracy as the evaluation metric to quantitatively assess classification performance. 

\noindent\textbf{Implementation Details.} 
The provided dataset includes original RGB videos and skeletal data extracted using OpenPose, featuring 137 full-body keypoints. 
To optimize data, we select 36 keypoints for the upper-body, facial landmarks, and hands. We also enhance data representation by generating additional modalities: depth using the method by \textit{Chen et al.}~\cite{chen2025video}, Taylor video modality via \textit{Wang et al.}'s~\cite{wang2024taylor} approximation, and optical flow through \textit{Dong et al.}'s~\cite{dong2024memflow} MemFlow approach.
For modeling, PoseConv3D~\cite{duan2022revisiting} is used to capture spatial-temporal dynamics in skeletal information (J), limb connections (L), and combined RGB with skeletal data (RGB+J and RGB+L). VideoSwin Transformer~\cite{liu2022video} is applied to RGB, depth, Taylor, and optical flow modalities for spatial-temporal processing. To enhance robustness, we perform transfer learning with VideoSwinT: initially pretraining on RGB data from Micro-Action 52 (MA-52)~\cite{guo2024benchmarking}, followed by fine-tuning on the iMiGUE dataset~\cite{liu2021imigue}.
Finally, we employ an ensemble fusion strategy, assigning weights to each modality based on contribution and correlation. We integrate \textbf{RGB*}, \textbf{Taylor}, \textbf{Flow}, and \textbf{Depth} from VideoSwin, along with \textbf{RGB+Joint} and \textbf{RGB+Limb} from PoseConv3D.

\subsection{Experimental Results}
We evaluated the proposed method on the iMiGUE dataset and compared its performance against state-of-the-art methods reported in the MiGA Challenges from 2023 to 2025. As presented in Table~\ref{tab:kaggle_results}, we provide the classification results of the top three competitors from these three consecutive editions, clearly demonstrating the consistent superiority of our proposed method over previous best-performing approaches across all years. Specifically, our approach achieved a Top-1 accuracy of 73.213\%, ranking first in the 2025 competition, significantly outperforming the second-place accuracy of 68.697\%. Compared with the best performance in the 2024 MiGA Challenge, our method realized an improvement of approximately 3\%, thus substantially exceeding the results from the 2023 edition as well.

\begin{table}[t!]
\caption{
Top-3 micro-gesture classification results from MiGA Challenges (2023–2025).  
Results are sourced from official competition leaderboards\protect\footnotemark[1] \protect\footnotemark[2] \protect\footnotemark[3].
$\mathbf{J}$ denotes the Joint modality;  
$\mathbf{L}$ denotes the Limb modality;  
$\mathbf{R}$ denotes the RGB video modality;  
$\mathbf{T}$ denotes the Taylor video modality;
$\mathbf{F}$ denotes the Optical Flow video modality;  
$\mathbf{D}$ denotes the Depth video modality.
}
\resizebox{1.0\linewidth}{!}{
\begin{tabular}{c|c|c|c|c}
\toprule
\textbf{Rank} & \textbf{Team} & \textbf{Backbone} & \textbf{Modality} & \textbf{Top-1 Acc (\%)} \\ 
\hline
MiGA'25 1st & gkdx2 (\textbf{Ours}) & PoseConv3D+VideoSwinT & $\mathbf{J+L+R+T+F+D}$ & \textbf{73.213}  \\
MiGA'25 2nd & awuniverse & - & - & 68.697 \\
MiGA'25 3rd & Lonelysheep & PoseConv3D & $\mathbf{J+L}$  & 67.010 \\
\hline
MiGA'24 1st & HFUT-VUT~\cite{chen2024prototype} & PoseConv3D & $\mathbf{J+L+R}$ & 70.254 \\
MiGA'24 2nd & NPU-MUCIS~\cite{huang2024multi} & Res2Net3D+GCN & $\mathbf{J+R}$ & 70.188 \\
MiGA'24 3rd & ywww11 & PoseConv3D+CLIP & $\mathbf{J+R}$ & 68.917 \\ 
\hline
MiGA'23 1st & HFUT-VUT~\cite{li2023joint} & PoseConv3D & $\mathbf{J+L}$ & 64.12 \\
MiGA'23 2nd & NPU-Stanford~\cite{huang2023micro} & Hyperformer & $\mathbf{J}$ & 63.02 \\
MiGA'23 3rd & ChenxiCui~\cite{xu2025towards} & - & - & 62.63 \\ 
\bottomrule
\end{tabular}
}
\label{tab:kaggle_results}
\end{table}
\footnotetext[1]{The 1st MiGA-IJCAI Challenge (2023) Track 1 Leaderboard: \href{https://codalab.lisn.upsaclay.fr/competitions/11758\#results}{https://codalab.lisn.upsaclay.fr/competitions/11758\#results}}
\footnotetext[2]{The 2nd MiGA-IJCAI Challenge (2024) Track 1 Leaderboard: \href{https://www.kaggle.com/competitions/2nd-miga-ijcai-challenge-track1/leaderboard}{https://www.kaggle.com/competitions/2nd-miga-ijcai-challenge-track1/leaderboard}}
\footnotetext[3]{The 3rd MiGA-IJCAI Challenge (2025) Track 1 Leaderboard: \href{https://www.kaggle.com/competitions/the-3rd-mi-ga-ijcai-challenge-track-1/leaderboard}{https://www.kaggle.com/competitions/the-3rd-mi-ga-ijcai-challenge-track-1/leaderboard}}

\begin{table}[t!]
\tabcolsep 5pt
\caption{Comparison of classification performance using different combinations of modalities and backbone architectures on the iMiGUE test set. The evaluated backbone models include PoseConv3D~\cite{duan2022revisiting} and VideoSwinT~\cite{liu2022video}. We evaluate six modalities: Joint, Limb, RGB, Taylor, Optical Flow, and Depth. Particularly, RGB* denotes that transfer learning was adopted by first pre-training on the Micro-Action 52 dataset~\cite{guo2024benchmarking} and subsequently fine-tuning on the iMiGUE dataset~\cite{liu2021imigue}.}
\resizebox{1.0\linewidth}{!}{
\begin{tabular}{c|ccc|cccc|c}
\toprule
\textbf{Backbone} & \textbf{Joint} & \textbf{Limb} & \textbf{RGB} & \textbf{RGB*} & \textbf{Taylor} & \textbf{Flow} & \textbf{Depth} & \textbf{Top-1 Acc (\%)} \\ 
\hline
PoseConv3D & \ding{51} & & & & & & & 65.256 \\
PoseConv3D & & \ding{51} & & & & & & 64.686 \\
PoseConv3D & & & \ding{51} & & & & & 64.511 \\
PoseConv3D & \ding{51} & \ding{51} & & & & & & 67.229 \\ 
PoseConv3D & \ding{51} & & \ding{51} & & & & & 68.917 \\
PoseConv3D & & \ding{51} & \ding{51} & & & & & 68.917 \\ 
\hline
VideoSwinT & & & \ding{51} & & & & & 65.629 \\
VideoSwinT & & & & \ding{51} & & & & 66.615 \\
VideoSwinT & & & & & \ding{51} & & & 62.845 \\
VideoSwinT & & & & & & \ding{51} & & 61.617 \\
VideoSwinT & & & & & & & \ding{51} & 65.212 \\
\hline
PoseConv3D+VideoSwinT & \ding{51} & & & \ding{51} & & & & 70.955 \\ 
PoseConv3D+VideoSwinT & & \ding{51} & & \ding{51} & & & & 70.802 \\ 
PoseConv3D+VideoSwinT & \ding{51} & \ding{51} & & \ding{51} & & & & 71.416 \\ 
PoseConv3D+VideoSwinT & \ding{51} & \ding{51} & & \ding{51} & \ding{51} & & & 72.095 \\ 
PoseConv3D+VideoSwinT & \ding{51} & \ding{51} & & \ding{51} & \ding{51} & \ding{51} & & 72.227 \\
PoseConv3D+VideoSwinT & \ding{51} & \ding{51} & & \ding{51} & \ding{51} & \ding{51} & \ding{51} & 72.644 \\
\hline
MM-Gesture (\textbf{Ours}) & \ding{51} & \ding{51} & \ding{51} & \ding{51} & \ding{51} & \ding{51} & \ding{51} & \textbf{73.213} \\
\bottomrule
\end{tabular}
}
\label{tab:modality_ensemble_results}
\end{table}

Here, we conduct comprehensive experimental settings to evaluate multiple modalities, including skeleton data (joints and limbs), RGB frames, Taylor series approximation videos (Taylor), optical flow, and depth information. As shown in Table~\ref{tab:modality_ensemble_results}, two backbone frameworks, namely PoseConv3D~\cite{duan2022revisiting} and VideoSwin~\cite{liu2022video}, were employed to thoroughly explore performance across various modality combinations. Experimental outcomes demonstrate that while single-modality inputs generally show moderate competitiveness, they nevertheless yield relatively lower accuracies, highlighting the inherent challenges of relying on a single modality in micro-gesture classification tasks. However, the incorporation of multiple modalities consistently results in enhanced performance, clearly emphasizing the complementary and distinctive nature of the various modalities in improving classification accuracy.

Our subsequent multimodal fusion experiments verify the complementary nature of diverse data streams. Specifically, integrating skeleton (joint and limb) data with RGB frames results in an accuracy improvement to 71.416\%, clearly demonstrating the strength of combining structural and appearance-based representations. Incorporating the Taylor modality further boosts accuracy to 72.096\%, reflecting benefits from pixel-level temporal-spatial approximations that effectively capture subtle dynamic gestures. Additional integration of optical flow and depth modalities improves performance even further, reaching an accuracy of 72.644\%, confirming their roles as valuable supplementary information sources. 
Ultimately, through an optimized multimodal fusion weighting strategy, our method achieves a Top-1 accuracy of 73.213\%. These results strongly affirm the advantages of properly designed multimodal fusion techniques and emphasize the efficacy and robustness of the presented approach over previously published state-of-the-art methods in micro-gesture recognition tasks.


\section{Conclusion}
In this paper, we proposed \textbf{MM-Gesture}, a novel multimodal ensemble framework for micro-gesture recognition. Our method integrates complementary features from six modalities—skeleton, limb, RGB, Taylor series approximation, optical flow, and depth—to leverage their distinct fine-grained characteristics. Additionally, we employed transfer learning by pretraining the RGB-based model on the Micro-Action 52 dataset before fine-tuning on the target iMiGUE dataset. Experiments demonstrate that our multimodal fusion significantly outperforms single or fewer modality baselines. Our model achieved a top-1 accuracy of 73.213\% on the challenging iMiGUE dataset, ranking first in the 3rd MiGA Competition at IJCAI 2025. 

For future work, we aim to explore the integration of Multimodal large language models (MLLMs)~\cite{xu2024mc,xu2024gg} and skeleton-based micro-gesture encoders. We plan to utilize MLLMs' rich semantic understanding and extensive prior knowledge to enhance micro-gesture recognition through interactive prompts and contextual reasoning, further advancing multimodal and affective human behavior understanding. Additionally, we will incorporate modalities such as gaze~\cite{liu2024depth}, audio~\cite{zhao2025temporal}, and remote photoplethysmography (rPPG)~\cite{qian2024cluster} to enable comprehensive multimodal emotion analysis.

\begin{acknowledgments}
This work was supported by the National Natural Science Foundation of China (62272144,72188101,62020106007, and U20A20183), the Major Project of Anhui Province (202203a05020011), the Fundamental Research Funds for the Central Universities (JZ2024HGTG0309), and the Earth System Big Data Platform of the School of Earth Sciences, Zhejiang University.
\end{acknowledgments}

\bibliography{sample-ceur}

\end{document}